\documentclass{article}

\usepackage[preprint]{neurips_2021}

\usepackage[utf8]{inputenc} % allow utf-8 input
\usepackage[T1]{fontenc}    % use 8-bit T1 fonts
\usepackage{hyperref}       % hyperlinks
\usepackage{url}            % simple URL typesetting
\usepackage{booktabs}       % professional-quality tables
\usepackage{amsfonts}       % blackboard math symbols
\usepackage{nicefrac}       % compact symbols for 1/2, etc.
\usepackage{microtype}      % microtypography
\usepackage{xcolor}         % colors
\usepackage{wrapfig}
\usepackage{bm}
\usepackage{amsmath}
\usepackage{amssymb}
\usepackage{amsthm}
\usepackage[inline]{enumitem}
\usepackage{algorithm}
\usepackage[noend]{algpseudocode}
\usepackage{graphicx}
\usepackage{diagbox}
\usepackage{multirow}
\usepackage{apptools}
\usepackage{diagbox}
\usepackage{makecell}
\usepackage{mathtools}
\usepackage{caption}
\usepackage{subcaption}

\DeclareMathOperator*{\argmax}{arg\max}
\newcommand\given[1][]{\:#1\vert\:}

\title{Reinforcement Learning to Solve NP-hard Problems: an Application to the CVRP}

\author{
Leo Ardon \\
Imperial College London
}

\begin{document}

\maketitle

\begin{abstract}
In this paper, we evaluate the use of Reinforcement Learning (RL) to solve a classic combinatorial optimization problem: the Capacitated Vehicle Routing Problem (CVRP). We formalize this problem in the RL framework and compare two of the most promising RL approaches with traditional solving techniques on a set of benchmark instances. We measure the different approaches with the quality of the solution returned and the time required to return it. We found that despite not returning the best solution, the RL approach has many advantages over traditional solvers. First, the versatility of the framework allows the resolution of more complex combinatorial problems. Moreover, instead of trying to solve a specific instance of the problem, the RL algorithm learns the skills required to solve the problem. The trained policy can then quasi instantly provide a solution to an unseen problem without having to solve it from scratch. Finally, the use of trained models makes the RL solver by far the fastest, and therefore make this approach more suited for commercial use where the user experience is paramount. Techniques like Knowledge Transfer can also be used to improve the training efficiency of the algorithm and help solve bigger and more complex problems.
\end{abstract}

\section{Introduction}

Combinatorial optimization problems can be found in various fields around us. From finance to logistics and supply chain planning, we can find variants of some of the standard optimization problems (such as the Knapsack problem, the Traveling Salesman Problem or TSP and the Vehicle Routing Problem or VRP). Despite having a rather simple goal: finding the optimal solution in a finite set of objects; solving this type of problems turns out to be challenging due to the exponentially increasing complexity to find a solution as the size of the problem grows. The complex nature of the problem and its high applicability to real-world problems stirred up the interest of the research community and led, throughout the years, to many algorithms proposed to solve the problems. Most of these are however problem specific and require solving the problem from scratch every time. Until recently, we could classify these techniques in two main families: the exact methods and the heuristic approach, but a new trend led by the Machine Learning community started to emerge. This new approach consists of learning heuristics with models rather than hand-written logic currently needed by existing methods. The recent progress in Reinforcement Learning (a sub-field of Machine Learning) and the prospect of finding new algorithms without the need of hand-crafted heuristics make this technique an obvious candidate to address the challenges combinatorial optimization problems are subject to: scalability and generality.

In this paper, we focus our attention to the Capacitated Vehicle Routing Problem known to be computationally difficult, that has been the subject of many research papers but for which finding a good solution in a reasonable amount of time for larger problems remains challenging. We compare the Reinforcement Learning approach with the more traditional techniques on a set of benchmark instances traditionally used by the operations research community and identify the advantages and drawbacks of this new technique to solve combinatorial optimization problems.

\subsection{Related Work}

Combinatorial Optimization problems and more particularly routing problems have been studied extensively in the operations research community. However, it is only until recently and the work from \cite{Bello_Pham_Le_Norouzi_Bengio_2017} that Reinforcement Learning was used to solve routing problems. They train a Neural Network called Pointer Network, initially introduced by \cite{Vinyals_Fortunato_Jaitly_2015} and used to solve the TSP, that autoregressively learns to create a solution to the problem by adding one node at a time to a partial tour using an attention mechanism to encode the state of the problem. \cite{Bello_Pham_Le_Norouzi_Bengio_2017} use a Policy Gradient algorithm to train the recurrent neural network with RL and demonstrate that without much engineering and problem specific heuristic design, combinatorial optimization problems can be solved. This approach, although promising, required intensive computing power. The training of the Pointer Network is slow and complex, and the time needed to obtain a satisfactory solution becomes prohibitive as the size of the problem increases. In addition, although it works on the TSP, the Pointer Network is unfortunately not applicable to more complex combinatorial optimization problems in which the state of the problem varies over time, such as the CVRP. The network would need to be updated every time the state changes making the algorithm even more resource demanding. \cite{Nazari_Oroojlooy_Snyder_Takac_2018} try to tackle this problem by simplifying the architecture of the network. They recognize that the Recurrent Neural Network used in the Pointer Network and often applied for sequence modelling, unnecessarily relies on the order of the sequence of the input. The order in which the location of each customer is passed as an input shouldn’t have any effect on the final solution. They propose a new model, invariant to the input sequence, that considerably decreases the computational complexity of the algorithm. One year later, \cite{Kool_vanHoof_Welling_2019} attack the problem from two angles. First, they try to use another RL technique called REINFORCE to train their model. They also introduce yet another model used to capture the state of the system. Their approach is an extension of the attention mechanism used in the Pointer Networks but leverage the Transformer architecture from \cite{Vaswani_Shazeer_Parmar_Uszkoreit_Jones_Gomez_Kaiser_Polosukhin_2017} and use multiple multi-head attention layers to allow a more flexible and accurate representation of the state of the environment. The new neural network architecture yields better performances with a reduced gap statistic. The results presented also show that the REINFORCE method improves the learning efficiency with the algorithm converging faster than the technique used previously \citep{Nazari_Oroojlooy_Snyder_Takac_2018}, opening the door to a more scalable approach to solve routing problems with Reinforcement Learning.

\subsection{Research Objectives}

Solving combinatorial optimization problems with the Reinforcement Learning (RL) framework is a fairly new trend compared to the decades of research and the hundreds of algorithms developed for exact and heuristics methods to solve this type of problem. The recent advancement in the field of RL and the increasing computing resources available, has unlocked many new possibilities including the hope to solve combinatorial optimization problems in a more generic fashion. The CVRP, easily explainable but also famously hard to solve at scale, was a perfect candidate to put the state-of-the-art RL algorithms to the test. On the other hand, in the Operations Research field researchers have been working on the CVRP for many years and a multitude of algorithms have been designed to solve it either optimally or heuristically. To thoroughly assess the efficiency of a new algorithm, the research community established a library of classical problems, known to be challenging, to serve as a benchmark and help researchers understand the superiority of a technique over one another.

Despite presenting promising results, the application of RL to solve the CVRP was assessed on randomly generated CVRP instances. We argue that this approach offers good insights on how well the model generalizes but does not provide an accurate evaluation of the model that can be contrasted with more traditional techniques. To our knowledge, a comparison between standard techniques and a RL-based approach was never performed on some of the classical benchmarks of the CVRP. In this paper we wish to answer the following research questions:
\begin{itemize}
    \item Using the CVRP as an example, can we use the Reinforcement Learning framework to solve a combinatorial optimization problem?
    \item How does a RL-based solution compare to the existing Combinatorial Optimization techniques used to solve the CVRP on classical benchmark problems?
    \item What are the benefits of using RL to solve the problem?
\end{itemize}

We start by presenting the formulation of the CVRP using the Mixed Integer Linear Problem formalism and give a brief overview of the traditional combinatorial optimization methods for both exact and heuristic methods. We then introduce the main concepts of Reinforcement Learning and then define the CVRP in the RL framework. Finally, through a set of computational experiments, we try to answer these questions, evaluating the pros and cons of a RL-based approach to solve the CVRP.

\section{Problem Definition}

One of the most studied combinatorial optimization problems is the Vehicle Routing Problem (VRP), where the goal is to determine the cost-minimizing routes a fleet of vehicles needs to follow to serve a given set of customers. Initially introduced by \cite{Dantzig_Ramser_1959} as a generalization of the famous Traveling Salesperson Problem (TSP), the routing problem has since then been declined in a multitude of variants. Despite being easy to state in layman terms, the VRP belongs to the family of NP-hard problems, characterized by their exponentially increasing complexity to find a solution as the problem expands. The VRP is therefore a challenging task that nevertheless has many applications in industry. To cite just a few, we can think of the e- commerce giant Amazon, delivering 2.5 billion packages a year \citep{Cheng_2019}, constantly trying to optimize every step of the process including finding the best route their drivers should take to deliver the parcels in time and at a minimum cost. The problem solved by the ride-sharing company Uber is also a variant of the VRP where the company wants to assign the best possible routes to drivers to pick up and drop off passengers. The complexity of the problem and its applicability to the real-world makes the VRP very popular amongst both academic and industrial research communities and has been the subject of extensive studies generating more than hundreds of research papers and algorithms in the past 60 years.

\begin{figure}[ht]
  \centering
  \includegraphics[scale=0.45]{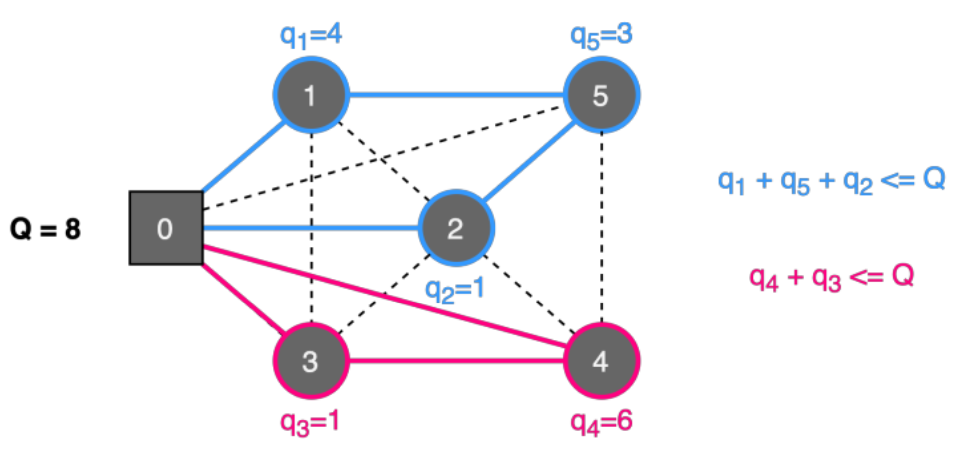}
  \caption{CVRP Example}
  \label{cvrp_example}
\end{figure}

In the context of our research, we focus our attention to the Capacitated Vehicle Routing Problem (CVRP), which constitutes one of the canonical variants of the VRP and is often used in the academic literature. In the CVRP, the goal is to satisfy the demand $q_i \geq 0$ of each of the $n$ customers $i \in {1,2,...,n}$, with a fleet of $K$ homogeneous vehicles, all starting from a given node, called the depot (often denoted as node 0), and with a fixed capacity $Q > 0$ of transported goods. A vehicle will visit customers one by one satisfying their demand in full and returning to the depot once it has delivered all its truckload. The cost of traveling between two nodes $i$ and $j$ is simply calculated as the Euclidean distance between the nodes: $c_{ij} = \sqrt{(x_i - x_j)^2 + (y_i - y_j)^2}$ where $(x_i, y_i)$represent the coordinates of the node $i$. The objective of the CVRP is to satisfy the demand of all the customers while minimizing the total distance travelled by the vehicles. The CVRP is often represented using a fully connected graph where the depot and the customers are materialized by vertices and the delivery truck is traveling along the edges and paying a cost $c_{ij}$ if the edge $i \rightarrow j$ is selected in the solution. The Figure \ref{cvrp_example} provides a simple example of the CVRP represented as a graph.

The squared node $0$ is the depot, the circle nodes are the clients requesting a quantity $q_i$ of the product being delivered, the graph is fully connected and undirected. The dotted lines represent the paths that were not selected in the proposed solution. The cost associated with each edge is the Euclidean distance between two nodes. The highlighted paths represent a feasible solution as all the $5$ customers have been visited and the sum of the quantities $q_i$ satisfied in each of the $2$ tours, is less or equal than the capacity $Q$ of a delivery truck. Note that this solution is also optimal as the total distance travelled is minimal.

We present below two mathematical models used to solve the CVRP using the two techniques we wish to compare in this paper. The first one, considered as the classical approach, is based on Mixed Integer Programming while the second method, more recent, makes use of Reinforcement Learning to solve the problem.

\subsection{Mixed Integer Programming}

The first paper formalizing the Vehicle Routing Problem \citep{Dantzig_Ramser_1959} also proposed a rigorous mathematical formulation and an algorithmic approach to solve the CVRP. Like most of the models currently existing, Mixed Integer Programming was used, defining an objective function and a set of constraints able to characterize the problem in full.

\subsubsection{Formulation}

From the literature, 4 main formulations of the CVRP stand out, each with their own advantages and drawbacks \citep{Toth_Vigo_2014}. We present below the two-index formulation for the undirected CVRP, also referred to as VRP2, and we use the MTZ-formulation \citep{Miller_Tucker_Zemlin_1960} to express the Subtour Elimination Constraints. This formulation will be used during the experiments phase with AMPL + Gurobi.

\begin{align*}
    &\min_e \sum^n_{i=0} \sum^n_{j=0} c_{ij}e_{ij} & \label{eq:1.1} \tag{1.1} \\
    &\sum^n_{j=0,i \neq j} e_{ij} = 1 &&\forall i \in \left\{1 \cdots n \right\} \label{eq:1.2} \tag{1.2} \\
    &\sum^n_{i=0,i \neq j} e_{ij} = 1 &&\forall j \in \left\{1 \cdots n \right\} \label{eq:1.3} \tag{1.3} \\
    &u_i - u_j + Q e_{ij} \leq Q - q_j &&\forall i,j \in \left\{1 \cdots n \right\} \label{eq:1.4} \tag{1.4} \\
    &u_i \geq q_i &&\forall i \in \left\{1 \cdots n \right\} \label{eq:1.5} \tag{1.5} \\
    &u_i \geq Q &&\forall i \in \left\{1 \cdots n \right\} \label{eq:1.6} \tag{1.6} \\
    &e_{ij} \in \left\{ 0, 1 \right\} &&\forall i,j \in \left\{1 \cdots n \right\} \label{eq:1.7} \tag{1.7} \\
    &u_i \in \mathbb{Z}_{+} &&\forall i \in \left\{1 \cdots n \right\} \label{eq:1.8} \tag{1.8}
\end{align*}
\setcounter{equation}{1}

The decision variables of the problem are the binary variables $e_{ij}$ indicating whether the edge connecting the nodes $i$ and $j$ should be included in a tour, and the variables $u_i$ used by the MTZ Subtour Elimination Constraints. The parameters are the demand $q_i$ of each customer, the capacity $Q$ a truck can hold and the distance $c_{ij}$ between two nodes pre-computed using the location $(x_i, y_i)$ of each node.

In order to minimize the distance travelled or traveling cost (eq. \ref{eq:1.1}), we can intuitively think that we should limit the back and forth between the nodes. Especially in the simple CVRP where the demand of the clients cannot be partially fulfilled, we can see the problem as trying to find elementary paths (that visit the customers only once). This is formalized with the constraints (eq. \ref{eq:1.2}) and (eq. \ref{eq:1.3}) making sure that a customer node has an out-degree and an in-degree of exactly 1, indicating that it should be visited only once for the solution to be feasible. The constraints (eq. \ref{eq:1.4}) to (eq. \ref{eq:1.6}) are used to control that a vehicle does not serve more demand than its capacity and to exclude sub-tours not including the depot. This formulation was first introduced by \cite{Miller_Tucker_Zemlin_1960} to solve the TSP and has been adapted to the CVRP. The variables $u_i$ represents the accumulated demand already satisfied by a vehicle before arriving in node $i$. The main advantage of this formulation is the tractability in the number of variables and constraints: $O(n^2)$. We note that we do not impose any constraint on the number of vehicles (or the number of tours) used to solve the problem as opposed to some other formulations \citep{Laporte_Nobert_Desrochers_1985}.

In the next sections, we summarize two families of algorithms used to solve this problem. The first family not only intends to find a solution but guarantees that the algorithm returns an optimal solution to the problem. On the other hand, the second group of algorithms, returns a “good-enough” solution, without any assurance on its optimality.

\subsubsection{Exact Algorithms}

We present in this section an overview of the main features of the exact algorithms developed to solve the CVRP. This type of algorithm aims at finding a solution with a guarantee of optimality. However, the NP-hard nature of the problem makes this condition difficult to satisfy as the problem size increases, the solution space becomes intractable and ensuring the optimality of the solution proposed, in a reasonable amount of time, becomes harder. Thanks to an active research community in the combinatorial optimization field, a range of new exact techniques emerged in the past 50 years. Branch-and-Bound followed by Branch-and-Price and Branch-and-Cut techniques were introduced, all of which declined in many algorithms that were applied to the VRP and showed great promise in the journey to find scalable methods to solve the problem. At the time, only limited computing resources were available, but these new techniques could solve problems involving up to 50 customers. \cite{Fukasawa_Longo_Lysgaard_Aragao_Reis_Uchoa_Werneck_2005} had the idea to combine the Branch-and-Price and Branch-and-Cut algorithms and demonstrated that these two complementary techniques put together could yield much better results than existing methods. With this new approach, instances involving up to 135 nodes could be solved to optimality. A detailed description of the algorithm, called Branch-and- Price-and-Cut, was later published by \cite{Desrosiers_Lubbecke_2011}.

In the context of the CVRP the Branch-and-Price-and-Cut has seen many implementations. Additional cuts were found to tighten the formulation of the problem and reduce the feasible region to explore: “Strong Degree Cuts” \citep{Contardo_Cordeau_Gendron_2011}, “Strengthened Capacity Cuts” \citep{Baldacci_Christofides_Mingozzi_2007}, ... Combined with these additional constraints, the pricing strategy was also revisited to improve the dual bound of the problem: \cite{Fukasawa_Longo_Lysgaard_Aragao_Reis_Uchoa_Werneck_2005} with the “$q$-routes without $k$-cycles” and \cite{Baldacci_Mingozzi_Roberti_2011} with the “$ng$-routes”. Finally, alternatives to the classical Branching were proposed; \cite{Baldacci_Christofides_Mingozzi_2007} enumerate all the elementary routes that may belong to an optimal solution and apply branch and cut to solve the set-partitioning problem with those routes. But the most efficient strategy to this day involves a hybrid strategy with route enumeration when tractable and branching otherwise \citep{Pecin_Pessoa_Poggi_Uchoa_2014}.

Despite seeing great improvement throughout the years in the ability to find optimal solutions for bigger and more complex problems, the time and the computational power needed to find a solution remain an important bottleneck, making this type of algorithms impractical to solve real-world problems.

\subsubsection{Heuristics Methods}

The guarantee of optimality is a strong requirement to meet which often requires a large amount of time to satisfy. As opposed to the exact approach, the heuristics method does not guarantee optimality but provides a reasonably good feasible solution in a more practical amount of time. The solution proposed using heuristics can in fact be very close to the optimal solution, sometimes within a couple of percent, and such solutions can be found for problems with a much bigger complexity that could not be solved via exact methods.

A basic heuristic method to solve combinatorial optimization problems consists of iteratively improving a randomly selected feasible solution via local search:
\begin{enumerate}
    \item We start by finding a feasible solution to the problem, that is a solution that need not be optimal but satisfy all the constraints.
    \item We then try to find an improved feasible solution by altering the current one with some transformations.
    \item If a better solution is found, we use it as our current solution for the next iteration starting from step 2. If no improved solution can be found, the current solution is a local optimum and the algorithm stops.
\end{enumerate}
The solver repeats this algorithm until the computation time limit is reached or a satisfactory solution is found.

This iterative improvement algorithm is the base of one of the most efficient heuristic methods used to solve the TSP \citep{Lin_Kernighan_1973}. In this paper we use the LKH3 solver, implementing the \cite{Lin_Kernighan_1973} heuristics efficiently for the CVRP \citep{Helsgaun_2017}. The basic idea of the algorithm is to iteratively apply the $\lambda$-opt algorithm (i.e exchanging $\lambda$ edges from the current solution to find a shorter tour) where the value of $\lambda$ is modified dynamically through the search to maximize the chance of finding a better solution.

The LKH3 has demonstrated impressive results on most of the classical benchmarks; it has found optimal solutions for most of the problems and has also helped improve some of the best-known solutions.

\subsection{Reinforcement Learning}

The recent breakthroughs in Reinforcement Learning (RL) and the promise to solve complex problems by only specifying an objective function to maximize \citep{Silver_Singh_Precup_Sutton_2021}, is gaining more and more attention in the research community. It was only a matter of time before the parallel between RL and Combinatorial Optimization was made. In this section, we start by introducing the main concepts of RL and use this formalism to model the CVRP.

\subsubsection{Reinforcement Learning Primer}

\begin{figure}[ht]
  \centering
  \includegraphics[scale=0.45]{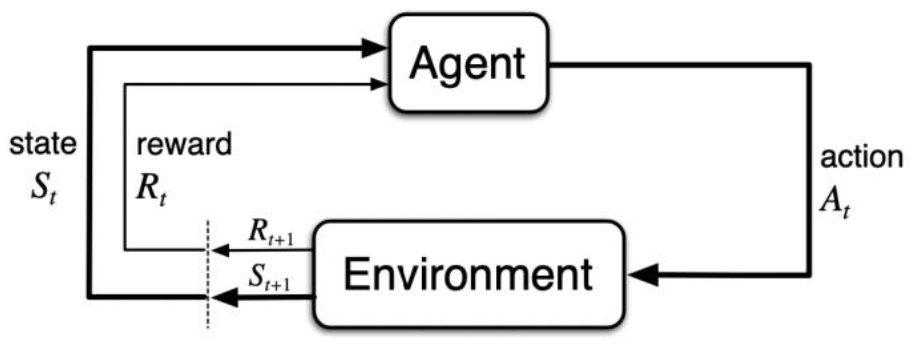}
  \caption{RL Interaction (Sutton \& Barto, 2018)}
  \label{environment_rl}
\end{figure}

Reinforcement Learning is a subfield of Machine Learning that cannot be classified under either the Supervised or the Unsupervised Learning paradigms. The principal characteristics of RL is the way the learning occurs; as opposed to the other families of learning techniques, reinforcement learning interacts with the environment to accumulate knowledge. By acting in the environment, the learning agent collects associated rewards to further infer the optimal strategy to adopt with the aim to maximize the cumulative rewards received. At every timestep, the agent decides which action $A_t$ to perform based on the current state of the environment $S_t$ and evaluate the “goodness” of this decision using the reward signal $R_t$ (Figure \ref{environment_rl})

Although not always satisfying the Markov Property, the problem of reinforcement learning is generally mathematically formalized as a Markov Decision Process (MDP) and is composed of:
\begin{itemize}
    \item $\mathcal{S}$: the set of states the environment can adopt.
    \item $\mathcal{A}$: the set of actions the learning agent can take.
    \item $\mathcal{T}[s, a, s']$: a transition function returning the probability to land in state $s'$ if the action $a$ was taken while in state $s$.
    \item $\mathcal{R}[s, a]$:a reward function returning the per-timestep reward of taking the action $a$ while in state $s$.
\end{itemize}

The solution of a MDP is called a policy: a function $a = \pi(s)$ returning an action $a$ to take while in state $s$. We define the value function $Q^{\pi}$ representing the long-term total expected discounted reward if we were to follow the policy $\pi$ starting in state $s$ and taking action $a$.

\begin{equation}\label{eq:2}
    Q^{\pi}(s, a) = \mathbb{E}_{\pi} \left[ \sum^{\infty}_{k=0} \gamma^k R_{t+k+1} \given[\Bigg] S_t = s, A_t = a \right]; \;\; \gamma \in \left[ 0, 1\right]
\end{equation}

The problem of reinforcement learning often involves estimating this value function by using the famous Bellman equation, mathematically expressing the recursive relationship satisfied by the value function. An optimal solution of the MDP $\pi^*$, called an optimal policy, can then be found by finding the optimal value function $Q^*$, and selecting at each step the action maximizing the future expected cumulative reward (eq. \ref{eq:3}), also known as the Bellman Optimality Equation.

\begin{equation}\label{eq:3}
    Q^*(s, a) = \mathbb{E}_{\pi^*}\left[ R_{t+1} + \gamma \max_{a' \in \mathcal{A}} Q^*(S_{t+1}, a') \given[\Bigg] S_t = s, A_t = a \right]
\end{equation}
\begin{equation}\label{eq:4}
    \pi^*(s) = \argmax_{a \in \mathcal{A}} Q^*(s, a)
\end{equation}

A method to solve the Bellman equation via iteration, consists of alternating Policy Evaluation and Policy Improvement. During the Policy Evaluation step the learning agent interacts with its environment and gradually updates the estimated value function $Q$ until convergence (eq. \ref{eq:3}). The second step leverages the value function that was previously learnt to improve the policy by taking greedy action; that is, for each state take the action that maximizes the expected cumulative reward (eq. \ref{eq:4}).

\begin{figure}[ht]
  \centering
  \includegraphics[scale=0.7]{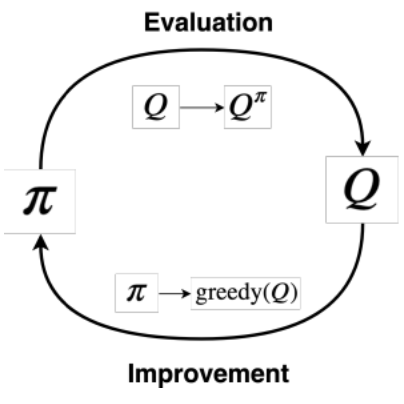}
  \caption{Value iteration}
  \label{value_iteration}
\end{figure}

As the number of states and actions increases, the number of values to store in memory explodes and the learning becomes very slow. To still be able to solve larger problems using RL, it is necessary to generalize the value functions using function approximation. A new set of parameters $\theta$ is introduced to allow a single function to support all the states the environment can adopt.

\begin{equation}\label{eq:5}
    Q_{\text{approx}}(s, a, \theta) = Q_{\theta}(s, a) \approx Q_{\pi}(s, a)
\end{equation}

The choice of the function approximators can vary depending on the problem at hand, but the generalization power of Neural Networks has made them the most popular choice nowadays and is often referred to as Deep-RL. The standard Machine Learning techniques such as Stochastic Gradient Descent, Quasi-Newton, ..., can be used to find the best parameter $\theta$ able to closely approximate the true value function, provided that the function approximator is differentiable.

The same method can be used directly on the policy function $\pi_{\theta}(s)$ and instead of learning the value function to then learn the policy, we can directly learn the policy function itself. This has the advantage of avoiding the extra step of “Evaluation” presented above. We define $J(\theta)$ (eq. \ref{eq:6}), a differentiable objective function measuring how good the policy is, that we wish to maximize and that will be used to drive the update of the parameter $\theta$. The Policy Gradient Theorem \citep{Sutton_Barto_2018} shows that we can express the update to apply to $\theta$ as a function of the value function $Q^{\pi_{\theta}}$ and the current policy $\pi_{\theta}$ (eq. \ref{eq:7}).

\begin{equation}\label{eq:6}
    J(\theta) = \sum_{s \in \mathcal{S}} d^{\pi}(s) \sum_{a \in \mathcal{A}} Q^{\pi}(s, a) \pi_{\theta}(a \given s)
\end{equation}
\begin{equation*}
    \text{where } d^{\pi} \text{is the state distribution of the system}
\end{equation*}

\begin{equation}\label{eq:7}
    \Delta \theta= \alpha \nabla_{\theta} J(\theta) = \alpha \mathbb{E}_{\pi_{\theta}} \left[ \nabla_{\theta}\log \pi_{\theta}(s \given a) Q^{\pi_{\theta}}(s, a) \right]
\end{equation}

The Policy Gradient Theorem is at the origin of many of the policy gradient algorithms proposed. A Monte-Carlo variant of the Policy Gradient Theorem called REINFORCE \citep{Williams_1992} was selected by \cite{Kool_vanHoof_Welling_2019} to try to solve the CVRP via RL. In this algorithm, the expected value is estimated via Monte-Carlo methods, running many iterations of the experiment to update the policy parameter $\theta$. However, the use of a Monte-Carlo based approach introduces high variance in the gradient of the REINFORCE algorithm. In fact, the stochasticity in the instances generated during training can lead to very different results in the total cumulative reward received at the end of each run and this high variance makes the learning process unstable and slow.

A popular solution to tackle this problem is to subtract a baseline from the approximated value $Q^{\pi_{\theta}}(s, a)$ to consider the variability across many experiments. A common baseline used in the literature is the state-value function expressing the expected discounted cumulative reward associated with a state $s$: $V_{\pi}(s) = \mathbb{E}_{\pi} \left[ \sum^{\infty}_{k=0} \gamma^k R_{t+k+1} \given S_t = s \right]$ leading to the following update in the gradient ascent:

\begin{equation}\label{eq:8}
    \Delta \theta = \alpha \mathbb{E}_{\pi_{\theta}} \left[ \nabla_{\theta} \log \pi_{\theta}(a \given s) A^{\pi_\theta}(s, a) \right]
\end{equation}
\begin{equation*}
    \text{with } A^{\pi_{\theta}}(s, a) = Q^{\pi_{\theta}}(s, a) - V^{\pi_\theta}(s) \text{ the Advantage function}
\end{equation*}

Now that the general RL methodology has been introduced, we return to its application to solve combinatorial optimization problems and more specifically the CVRP. In the next section, we formalize the problem in the MDP framework required by RL.

\subsubsection{Reinforcement Learning Formulation}

The CVRP can be seen as a sequential decision-making problem where at each timestep, the vehicle needs to decide where to go next to fulfil its task of serving all the customers. With this angle it becomes possible to express the components defining the RL problem: $<\mathcal{S}, \mathcal{A}, \mathcal{R}>$ the tuple of State space, Action space and Reward.
As we have previously mentioned, the CVRP can be represented as a graph where the nodes represent the customers to serve plus the depot. The current demand as well as the location of each customer is associated with their node on the graph. Every time a vehicle visits a node, either to satisfy the demand of a client or to reload the truck at the depot, the graph can be updated to reflect the current state of the system. The ensemble $\mathcal{S}$, composed of all the possible states $s$ the MDP can adopt, can therefore be encoded with the following values characterizing the graph:
\begin{itemize}
    \item $(x_0, y_0)$: The location of the depot
    \item $(x_i, y_i)$: The location of each customer $i$
    \item $k_t$: The current node the truck is visiting at time $t$
    \item $q_{it}$: The current demand of the customer at node $i$ at time $t$
    \item $Q_t$: The current load in the truck at time $t$
\end{itemize}

The location of the depot is important to know, especially when the truck is nearly empty, as it will need to go back to the depot to reload. Depending on the amount of goods left in the truck, it might be preferable to find a node closer to the depot as we know that the truck will soon need to go there. The current load in the truck is therefore also required to be included in the state space. The location of each node and the current location of the truck are important information to provide to the learning agent to understand the cost associated with traveling from one node to another. Finally, the current demand of each customer helps determine which nodes have not yet been visited and are candidates for the next location of the truck. More information can be added to the state space like the distance travelled, or the previous location. The more information that can help the neural network of the policy make the best decision the better. However, there is a trade-off to make as with too much information, we will face the “curse of dimensionality” and the training will likely take too long.

The action space $\mathcal{A}$ is the finite set of actions selecting which node the truck will visit next:
\begin{itemize}
    \item $\mathcal{A} = \left\{ 0 \cdots n \right\}$
\end{itemize}

In theory, the policy can select any node at any given timestep $t$. However, visiting the same node twice or traveling to the depot despite having enough load to satisfy at least one other customer is highly inefficient. Both \cite{Nazari_Oroojlooy_Snyder_Takac_2018} and \cite{Kool_vanHoof_Welling_2019} use a trick to guide the decision-making process:

\begin{itemize}
    \item When the current load of the truck cannot satisfy the demand of the customers, the corresponding nodes are masked out during the action selection process making sure that these nodes are not selected.
    \item When a customer node has its demand fulfilled, it is masked out until the end of the run.
\end{itemize}

This additional logic helps the exploration process of the RL algorithm and remains generic to any instances of the CVRP.

The reward signal $R$ of the problem is analogous to the MIP formulation: the goal being to find the cost-minimizing routes travelled by the truck, and the objective of the RL being to maximize the accumulated reward, we use the opposite of the total distance travelled.
\begin{itemize}
    \item $R = - \sum_{ij} c_{ij}x_{ij}$
\end{itemize}

The RL algorithm is exposed to many configurations of the problem with a fixed set of nodes, being in different locations each time. Exposing the learning agent to many instances helps the learnt policy to generalize and avoid overfitting to any specific CVRP instance. A run starts with all the customers’ demand unsatisfied and the truck at the depot, the agent interacts with the environment by visiting nodes until all the customers have had their demand fulfilled. The routes selected are only evaluated at the end of each episode producing the reward for this run.

The solution returned by the trained policy is guaranteed to be feasible thanks to the masking trick only allowing certain nodes to be selected. However, the solution proposed has no guarantee to be optimal, the algorithm learns to minimize the distance travelled but has no sense of the minimum that could be achieved, the exploration process of the algorithm could converge towards the optimal solution or be stuck in a local minimum. Therefore, the RL approach can only be classified as a heuristic method.

The CVRP can then be defined in the RL framework, leveraging the latest breakthroughs in the field to solve the problem in a more generic and timely fashion. We will then analyze how this technique compares with the existing and more traditional ways to solve the problem on some of the classical benchmarks from the operations research literature.

\section{Experiments}

\subsection{Dataset and Metrics}

With the years of research on the CVRP, people have proposed particularly challenging instances for which an optimal solution could not be found. Although some of them
have now been solved, the problems are still used as benchmarks to evaluate the quality of an algorithm. Because of time and computational constraints, the experiments presented in this paper were run on a subset of the challenging set of instances introduced by \cite{Christofides_Mingozzi_Toth_1979} namely the CMT1 (50 customers), CMT2 (75 customers), CMT3 (100 customers) and CMT11 (120 customers), for which a graphical representation of these instances are presented in the Appendix. The full definition of these instances, including the location of each node and the demand for each client, is available in the Capacitated Vehicle Routing Problem Library: CVRPLib \citep{Uchoa_2021}. Despite having been introduced to the research community in 1979, these instances of the CVRP were only solved optimally 20 years later by \cite{Cordeau_Laporte_Mercier_2001}. We believe that these instances represent a good baseline to run our comparison between the RL-based and the more traditional methods. The sizes of the instances are reasonable enough to be run on a basic machine while still being challenging.

We compare all the techniques by evaluating the quality of the solution found compared to the known optimal solution (eq. \ref{eq:9}). We also report the time taken to return a solution for each of the techniques used. In the context of this paper, we are mainly interested in fast solvers, returning a reasonably good solution in a short amount of time. Whenever possible, we set the computing time limit to 60 seconds, forcing the solver to return its current best solution when this time is reached. This threshold was set to simulate the maximum amount of time a random application user would be willing to wait to get an answer from their query before becoming impatient and abandon the tool.

\begin{equation}\label{eq:9}
    gap = \frac{proposed - optimal}{optimal}
\end{equation}

The Reinforcement Learning algorithm, like most of the Machine Learning, is composed of a training phase and a prediction phase. The time reported in the results is the time taken by the RL algorithm to predict the solution of the problem, the training time will be presented separately.

\subsection{Configurations}

The comparison is performed using five different solvers: one using exact method, two using heuristics and two using RL.

We use the software AMPL \citep{Fourer_Gay_Kernighan_1989} coupled with the Gurobi solver \citep{GurobiOptimization} to benchmark the exact method. The formulation presented in (eq. \ref{eq:1.1}) - (eq. \ref{eq:1.8}) is used and we set a computation time limit of 60 seconds to return the current best solution (if an optimal solution is found sooner it is returned directly and does not wait for the 60s to be over). The number of clients, the capacity of the truck, the cost matrix and the demand for each client are passed as parameters.

Two commonly used heuristics solvers are tested with the four CVRP instances: the state-of-the-art solver LKH3, specifically implemented to solve Routing Problems \citep{Helsgaun_2017} and a more generic combinatorial optimization solver developed by Google called OR-Tools \citep{Perron_Furnon_2019}. The OR-Tools library is configured to use the “SAVINGS” strategy, developed by \cite{Clarke_Wright_1964}, to find the initial feasible solution that will be iteratively improved. We use the recommended “GUIDED\_LOCAL\_SEARCH” as the metaheuristic to drive the search. Finally, we set the time limit to 60 seconds to return the best solution possible.

To evaluate the RL-based method we use the two main algorithms that have been developed to solve the CVRP and that showed the most promising results. The first one, developed by \cite{Nazari_Oroojlooy_Snyder_Takac_2018} uses a Recurrent Neural Network (RNN) encoder-decoder coupled with an attention mechanism to keep track of the state of the MDP changing over time. We run the algorithm with the default configuration:
\begin{itemize}
    \item embedding dimension = 128; used to encode the state
    \item RNN layer = 1
    \item hidden dimension = 128
    \item dropout = 0.1
    \item learning rate of 0.0001 for both the actor and the critic
    \item beam search width = 10
\end{itemize}

The code was modified to forbid partial delivery to a customer ensuring that each node is visited only once. A bug has also been fixed to use the maximum demand specified in the configuration as opposed to the hard coded value of $9$.

The second RL-based method evaluated comes from \cite{Kool_vanHoof_Welling_2019}. They use the REINFORCE method presented above. We ran the algorithm with the following configuration:
\begin{itemize}
    \item embedding dimension = 128
    \item hidden dimension = 128
    \item RNN layers = 3
    \item learning rate = 0.0001
    \item baseline = “rollout”
    \item beam search width = 5000
\end{itemize}

The RL algorithm did not allow the specification of a time limit to solve the problem. After investigation, it was found that most of the time spent by the RL algorithms was associated with the encoding of the current state of the graph.
We acknowledge that a more thorough hyper parametrization of the different models could have been done in order to get the best results possible. Here we assumed that the default values provided correspond to the most sensible choices working generically with most of the problems.

The experiments were run on a personal MacBook Pro using 1 GPU.

\subsection{Results}

\subsubsection{Benchmark Experiments}

We present in Table \ref{tab:benchmark_results} the results of the five different techniques used to solve the four instances selected. The variable $n$ indicates the number of clients in the instance while $Q$ is the capacity of the truck. The first column is the mean total distance obtained over 10 runs, the gap metric is presented in percentage and the time to provide a solution in second. Time marked with an asterisk indicates that the solver was stopped early as it reached the time limit.

It has been observed that for problems with more than 20 customers the exact methods usually do not return an optimal solution in reasonable times. It is therefore not surprising to see that the exact solver does not find the optimal solution under the time limit of 60 seconds We see however that within a minute the quality of the solution returned remains stable with a gap around 10\% irrespective of the size of the problem. The OR-Tools library, despite using a heuristic approach, fails to return with a satisfying solution within the time limit. On the other hand, the specialized solver LKH3 is by far the best solver if we consider only the quality of the solution returned. It found a nearly optimal solution for all the four instances evaluated. The time taken to find the solution is also very acceptable, with only 14 seconds for 100 customers and 39 seconds for 120.

\begin{table}[ht]
\centering
\caption{Solver Comparison Results}
\resizebox{\textwidth}{!}{%
\begin{tabular}{cc||cccccccccccc|}
\cline{3-14}
\multicolumn{1}{c}{} & \multicolumn{1}{c|}{} & \multicolumn{12}{c|}{Problem} \\ \hline
\multicolumn{2}{|c|}{\multirow{2}{*}{Method}} & \multicolumn{3}{c|}{\begin{tabular}[c]{@{}c@{}}CMT1\\ n = 50; Q = 160\end{tabular}} & \multicolumn{3}{c|}{\begin{tabular}[c]{@{}c@{}}CMT2\\ n = 75; Q = 140\end{tabular}} & \multicolumn{3}{c|}{\begin{tabular}[c]{@{}c@{}}CMT3\\ n = 100; Q = 200\end{tabular}} & \multicolumn{3}{c|}{\begin{tabular}[c]{@{}c@{}}CMT11\\ n = 120; Q = 200\end{tabular}} \\
\multicolumn{2}{|c|}{} & \multicolumn{1}{c}{Mean} & \multicolumn{1}{c}{\begin{tabular}[c]{@{}c@{}}Gap\\ (\%)\end{tabular}} & \multicolumn{1}{c|}{\begin{tabular}[c]{@{}c@{}}Time\\ (s)\end{tabular}} & \multicolumn{1}{c}{Mean} & \multicolumn{1}{c}{\begin{tabular}[c]{@{}c@{}}Gap\\ (\%)\end{tabular}} & \multicolumn{1}{c|}{\begin{tabular}[c]{@{}c@{}}Time\\ (s)\end{tabular}} & \multicolumn{1}{c}{Mean} & \multicolumn{1}{c}{\begin{tabular}[c]{@{}c@{}}Gap\\ (\%)\end{tabular}} & \multicolumn{1}{c|}{\begin{tabular}[c]{@{}c@{}}Time\\ (s)\end{tabular}} & \multicolumn{1}{c}{Mean} & \multicolumn{1}{c}{\begin{tabular}[c]{@{}c@{}}Gap\\ (\%)\end{tabular}} & \multicolumn{1}{c|}{\begin{tabular}[c]{@{}c@{}}Time\\ (s)\end{tabular}} \\ \hline \hline
\multicolumn{1}{|c}{\begin{tabular}[c]{@{}c@{}}Optimal \\ Solution\end{tabular}} & CVRP Lib & 524.6 & 0.0 & \multicolumn{1}{c|}{?} & 835.2 & 0.0 & \multicolumn{1}{c|}{?} & 826.1 & 0.0 & \multicolumn{1}{c|}{?} & 1042.1 & 0.0 & ? \\ \cline{1-2}
\multicolumn{1}{|c}{\begin{tabular}[c]{@{}c@{}}Exact \\ Solver\end{tabular}} & \begin{tabular}[c]{@{}c@{}}AMPL + \\ Gurobi\end{tabular} & 548.4 & 11.4 & \multicolumn{1}{c|}{60*} & 888.7 & 6.4 & \multicolumn{1}{c|}{60*} & 924.9 & 12.0 & \multicolumn{1}{c|}{60*} & 1108.1 & 6.3 & 60* \\ \cline{1-2}
\multicolumn{1}{|c}{\multirow{2}{*}{Heuristics}} & OR-Tools & 556.5 & 6.1 & \multicolumn{1}{c|}{60*} & 890.8 & 6.6 & \multicolumn{1}{c|}{60*} & 875.1 & 5.9 & \multicolumn{1}{c|}{60*} & 1178.1 & 13.0 & 60* \\
\multicolumn{1}{|c}{} & LKH3 & 524.6 & 0.0 & \multicolumn{1}{c|}{12.7} & 836.2 & 0.1 & \multicolumn{1}{c|}{48.1} & 829.4 & 0.4 & \multicolumn{1}{c|}{14.1} & 1042.1 & 0.0 & 39.0 \\ \cline{1-2}
\multicolumn{1}{|c}{\multirow{2}{*}{RL}} & Nazari & 562.1 & 7.1 & \multicolumn{1}{c|}{7.2} & 907.5 & 8.6 & \multicolumn{1}{c|}{11.8} & 915.3 & 10.8 & \multicolumn{1}{c|}{17.5} & 1221.8 & 17.2 & 21.8 \\
\multicolumn{1}{|c}{} & Kool & 531.9 & 1.4 & \multicolumn{1}{c|}{1.5} & 867.2 & 3.8 & \multicolumn{1}{c|}{1.7} & 882.5 & 6.8 & \multicolumn{1}{c|}{1.9} & 1207.7 & 15.9 & 3.0 \\ \hline
\end{tabular}%
}
\label{tab:benchmark_results}
\end{table}

The RL-based solution using \cite{Kool_vanHoof_Welling_2019} provides reasonable solutions with for example a gap of 1.4\% for CMT1 and 3.8\% for CMT2. However, we note that as the problem size increases, the quality of the solution deteriorates: CMT11 has a gap of 15.9\%. This can be attributed to the increasing complexity to train the neural network encoder used to interpret the state of the graph needed to take the optimal action. As the size of the graph grows, the training of the encoder becomes more complicated. We believe that with potentially more training and hyper parametrization, better results could be obtained. Looking at the time taken by this algorithm however, we clearly see the advantage of a learnt policy over a classic solver: the learning being handled separately, the prediction is fast, and all the solutions are returned within 3 seconds.

The results using the algorithm from \cite{Nazari_Oroojlooy_Snyder_Takac_2018} are not as impressive, but we still see an improvement in the computation time of the solution compared to the classical methods.

\subsubsection{Overfitting Experiment}

As opposed to traditional methods, RL intends to learn the skills required to solve the problem more holistically rather than trying to find an optimal solution for one specific configuration of the problem. Where the other combinatorial optimization solving methods will need to solve the problem from scratch every time, the RL approach leverages what it has learnt before and the skills it has acquired to quickly provide a solution to a new problem. This generic approach comes at a cost though; to be able to generalize well, the learning agent needs to be exposed to many different configurations of the problem to learn how to solve it efficiently. This is done during the training phase which can be quite long and computationally expensive. In our experiments, both RL-based algorithms were trained for several days to expose the graph encoder to many different instances of the problem and learn to decide on the best node to visit next. It took 3 days to train the policy used to solve CMT1 with \cite{Kool_vanHoof_Welling_2019}’s algorithm over 100 epochs and 6 days with \cite{Nazari_Oroojlooy_Snyder_Takac_2018} on a single machine one training at a time. The time taken to train the model increased with the size of the problem.

To truly compare apples to apples, we tweak the RL learning strategy to learn to solve only one specific instance of the problem. We run an experiment where the RL algorithm is exposed many times to the CMT1 and try to learn the skills to solve this problem in particular. In the machine learning world, this approach can be seen as overfitting where the algorithm specializes at solving a specific input very well. This is generally bad practice as we never know what future data will look like, but in the context of this experiment we are trying to reduce the RL-based approach to the same conditions than the current combinatorial optimization methods.

\begin{figure}[ht]
  \centering
  \includegraphics[scale=0.5]{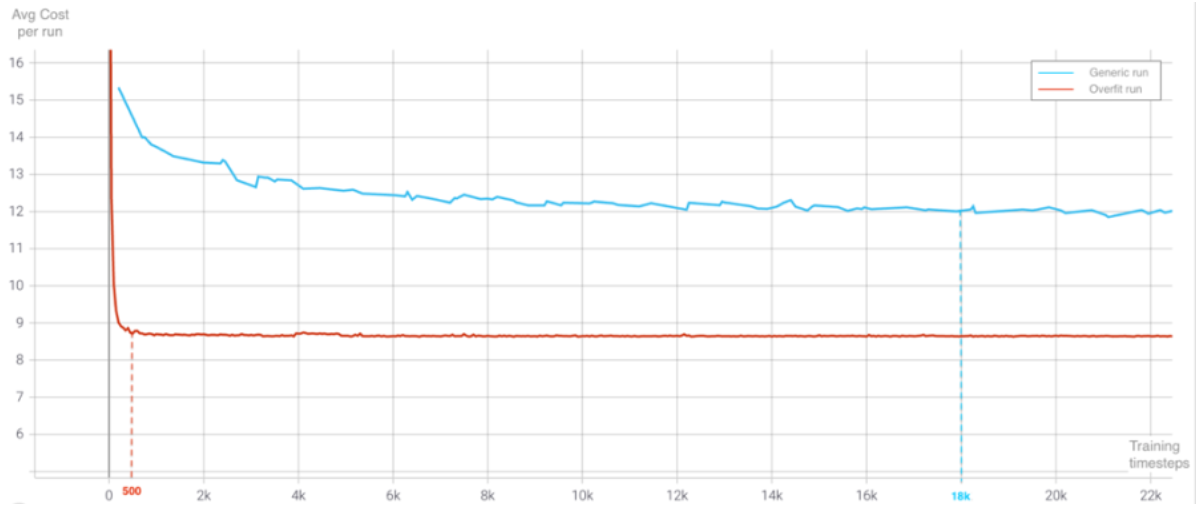}
  \caption{Overfit training average cost}
  \label{overfitting_experiment}
\end{figure}

The plot in Figure \ref{overfitting_experiment} represents the average distance travelled per run during the training phase, as the algorithm learn to optimize the routes to take. We compare the training of the overfitted policy with the generic policy initially learnt in the first set of experiments. We see that because the problem space is much larger for the generic policy, the algorithm requires much more time to converge. On the other hand, with only one problem to solve, the overfitted policy reaches an equilibrium 36 times faster. The difference in the values at which the algorithm converges can be explained by the fact that the generic policy is exposed to a multitude of instances, among which some require a larger distance to be travelled in order to be solved, causing the average cost to be higher. Looking at the results in Table \ref{tab:overfitting_results}, we see a small deterioration in the performance of the algorithm with a gap statistic of 2.6\% compared to 1.4\% obtained previously. In this experiment we then conclude that the overfitting helps speed up the training phase, however the quality of the model diminishes. In fact, by being exposed to different instances of the problem the neural network of the generic policy learns better to identify the most important features of the encoded state of the graph and will be better equipped to find a better route for the CVRP.

\begin{table}[ht]
\centering
\caption{Overfit Results}
\begin{tabular}{|cc||ccc|}
\hline
\multicolumn{2}{|c|}{\multirow{2}{*}{Method}} & \multicolumn{3}{c|}{\begin{tabular}[c]{@{}c@{}}CMT1\\ n = 50; Q = 160\end{tabular}} \\
\multicolumn{2}{|c|}{} & Mean & \begin{tabular}[c]{@{}c@{}}Gap\\ (\%)\end{tabular} & \begin{tabular}[c]{@{}c@{}}Time\\ (s)\end{tabular} \\ \hline \hline
\begin{tabular}[c]{@{}c@{}}Optimal \\ Solution\end{tabular} & CVRP Lib & 524.6 & 0.0 & ? \\ \hline
\multirow{2}{*}{RL} & Kool & 531.9 & 1.4 & 1.5 \\
 & \begin{tabular}[c]{@{}c@{}}Kool overfit\end{tabular} & 538.4 & 2.6 & 1.6 \\ \hline
\end{tabular}
\label{tab:overfitting_results}
\end{table}

\clearpage

The biggest drawback of the RL approach is also what makes its strength. Even though, the learning phase can be quite long to obtain satisfactory results, this phase is important to (i) be able to solve instantly a large variety of problems of the same structure (i.e., same number of nodes and same vehicle capacity) irrespective of their size, which is not the case using existing method and (ii) allow the algorithm to learn important latent features of the encoded state of the graph to yield better results.

\begin{figure}[ht]
  \centering
  \includegraphics[scale=0.5]{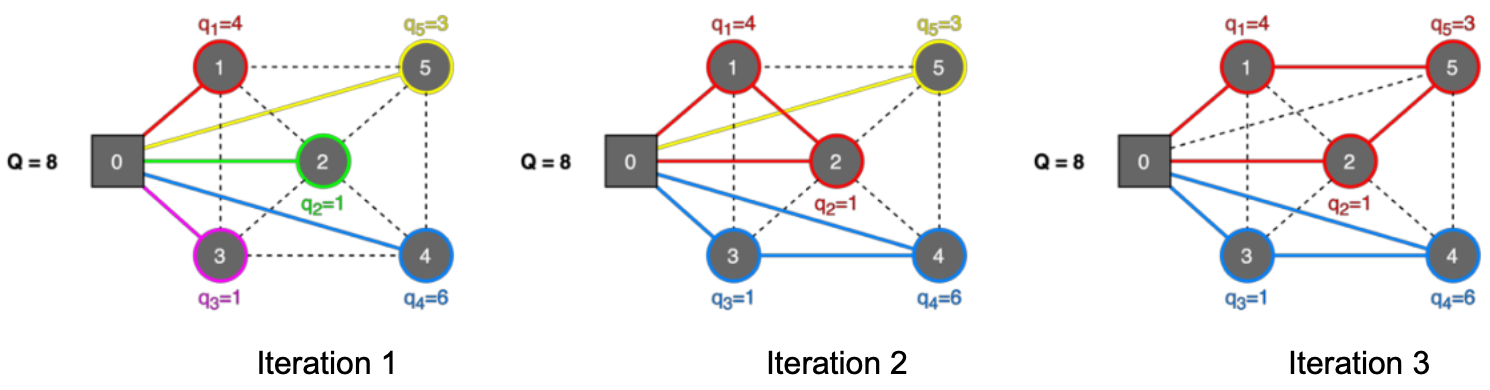}
  \caption{Replication of the Clark \& Wright heuristics}
  \label{rl_heuristics}
\end{figure}

Looking closer at the solution returned by the learning RL agent during the early phase of the training, we can draw a parallel with one of the earliest methods proposed to solve the CVRP. We see that at first, the agent creates one route for each customer, i.e., the vehicle travels from the depot, satisfies the demand of one customer, and comes back to the depot. As the algorithm observes that this solution, although feasible, does not yield the best reward, it learns to combine routes together to reduce the distance travelled (Figure \ref{rl_heuristics}). We cannot ignore the similarities with the heuristic approach presented by \cite{Clarke_Wright_1964} in 1964. The learning agent has learnt by itself one of the most popular heuristic method. We can think that potentially other known heuristics are learnt by the algorithm as it performs trial and error to find the best possible routes. The agent could also be learning heuristics that have not been considered yet by the research community but a deeper analysis of the evolution of the solutions proposed would be required and could be the object of future research.

\subsubsection{Reuse Experiment}

An exciting area of Machine Learning that has been growing in popularity is Transfer Learning, or the ability to reuse the knowledge previously acquired to accelerate the learning of a new task. Instead of solving the problem from scratch, previously learnt models or policies solving a similar problem can be reused.

\begin{figure}[ht]
  \centering
  \includegraphics[scale=0.5]{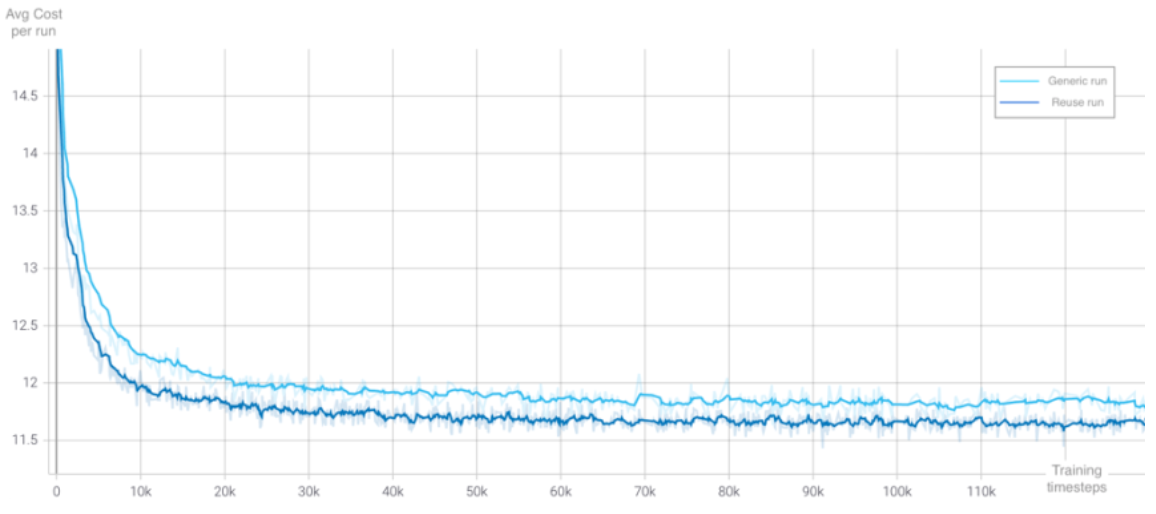}
  \caption{Reuse training average cost}
  \label{reuse_experiment}
\end{figure}

Solving the CVRP is notoriously hard, however the Traveling Salesperson Problem, a simplified version of the CVRP that can be solved more easily, shares many aspects with the CVRP. In fact, the CVRP can be seen as multiple TSPs. We leverage this approach by reusing a pre-trained TSP policy to train our algorithm to solve the CVRP. The goal of this experiment is to prove that skills that have been learnt previously from a simpler but similar problem can help speed up the learning of a new model. The features of the TSP, like the routing from one node to its neighbour, are then reused instead of being learnt from scratch by the algorithm. We modify the code from \cite{Kool_vanHoof_Welling_2019} to allow the reuse of different policies and present the difference in learning. We run an experiment using the CMT1 problem and evaluate the training phase of the algorithm compared to the previous run without reuse.

As we can see in Figure \ref{reuse_experiment} the use of a pre-trained TSP policy helps the models to learn faster and better with a lower average cost per run. When tested on the CMT1, the reuse policy returns the exact same solution than the generic one with a gap of 1.3\% (Table \ref{tab:reuse_results}) indicating that this approach helped speed up the training while not affecting the quality of the solution on CMT1. This technique tackles one of the main problem traditional combinatorial optimizations solving methods are facing, whereby the algorithm must solve the problem from the ground up over and over. Transfer learning could be the solution to the expensive computational requirements required to solve big instances of the CVRP and could also help solving more complex variants of the Vehicle Routing Problem.

\begin{table}[H]
\centering
\caption{TSP Reuse Results}
\begin{tabular}{|cc|ccc|}
\hline
\multicolumn{2}{|c|}{\multirow{2}{*}{Method}} & \multicolumn{3}{c|}{\begin{tabular}[c]{@{}c@{}}CMT1\\ n = 50; Q = 160\end{tabular}} \\
\multicolumn{2}{|c|}{} & Mean & \begin{tabular}[c]{@{}c@{}}Gap\\ (\%)\end{tabular} & \begin{tabular}[c]{@{}c@{}}Time\\ (s)\end{tabular} \\ \hline
\begin{tabular}[c]{@{}c@{}}Optimal \\ Solution\end{tabular} & CVRP Lib & 524.6 & 0.0 & ? \\ \hline
\multirow{2}{*}{RL} & Kool & 531.9 & 1.4 & 1.5 \\
 & \begin{tabular}[c]{@{}c@{}}Kool \\ TSP reuse\end{tabular} & 531.9 & 1.4 & 1.3 \\ \hline
\end{tabular}%
\label{tab:reuse_results}
\end{table}

\section{Conclusion}

In this paper we showed that Reinforcement Learning can be used to solve combinatorial optimizations problems and evaluated how this new approach compares with more traditional techniques. Through a set of experiments, we have identified some of the advantages and drawbacks of this technique to solve one of the most studied problems in the field: the CVRP.

We often refer to two families of algorithms used to solve combinatorial optimization problems, the exact and the heuristics methods. These two approaches serve different purposes and have different applications. One guarantees optimality and therefore takes longer to return a solution while the other returns a reasonably good solution in a timely manner. The Reinforcement Learning approach does not intend to solve the problem optimally and could therefore be considered as a heuristic approach. The similarities are such that we have observed the learning agent learnt by itself a heuristic developed by researcher in the 60s. However, it exists some fundamental differences between a RL-based and a heuristic algorithm. First, Reinforcement Learning requires a training phase where the algorithm, exposed to many different configurations, learns the best course of actions to take to maximize its objective. As we have seen in the experiments, this phase can be quite long and computer-intensive, but it also allows the model to generalize well. As opposed to heuristic methods, the same model can find a solution to a new unseen problem in a very short amount of time without having to re- train or solve the problem from scratch. Nonetheless, the training time can sometimes be prohibitive, especially with large instances of the problem, but tricks like the knowledge transfer technique presented in this paper can help reduced the time required to train the model. Based on the literature review, it seems that the improvements in the performance of RL algorithms, to reduce the training time and increase the quality of the solution returned, have mainly been driven by a better ability to encode the state of the graph and capture the important features that will help the decision-making. The recent development of the Graph Neural Network architecture, potentially more suited for this type of graphical problem than classic RNN, could be the object of future research. Another advantage of the Reinforcement Learning approach is the simplicity of the problem definition. We only need to define the rules of the problem and the algorithm learns by itself via trial and error how to solve it. This allows the resolution of much more complex problems, difficult to express via MILP (e.g. problems with stochasticity in the demand, in the routing cost, with time window, ...). From a performance point of view, the RL approach falls short of the specialized heuristic solver LKH3 able to return a near-optimal solution in less than one minute, but RL returns a reasonably good solution more than ten times faster.

We can then conclude, that thanks to its very fast query time and its ability to solve a wide range of problems instantly with the same model, the RL-based method answers a different set of requirements and could even be the object of a new combinatorial optimization solving technique family. An interesting avenue for future research could be the combination of both RL and heuristic approaches, to take advantage of the speed of RL to find a reasonably good initial solution and the power of heuristics to improve it, close to optimality.

\section*{Acknowledgement}
I would like to thank my supervisor, Dr. Wolfram Wiesemann, whose help and guidance was invaluable to help me drive this research project in the right direction. I appreciated his advice and the freedom I was given during our collaboration.

\clearpage

\bibliographystyle{apalike}
\bibliography{cvrp}

\end{document}